\documentclass{article}

    \usepackage[preprint]{neurips_2021}

\usepackage[utf8]{inputenc} 
\usepackage[T1]{fontenc}    
\usepackage{hyperref}       
\usepackage{url}            
\usepackage{booktabs}       
\usepackage{amsfonts, amsmath}       
\usepackage{nicefrac}       
\usepackage{microtype}      
\usepackage{acronym}
\usepackage{graphicx}
\usepackage{caption}
\usepackage{subcaption}
\usepackage{color, soul}
\usepackage{bm}
\usepackage[nocompress]{cite}

\acrodef{pan}[PAN]{path-integral-based graph neural networks}
\acrodef{grw}[GRW]{generic random walk}
\acrodef{met}[MET]{maximal entropy transition}
\acrodef{hpan}[HPAN]{heterogeneous PAN}
\acrodef{panc}[PANConv]{PAN convolution module}
\acrodef{panp}[PANPool]{PAN pooling module}
\acrodef{gnn}[GNN]{Graph neural network}

\title{Path Integral Based Convolution and Pooling for Heterogeneous Graph Neural Networks}

\author{
Lingjie Kong \\
Department of Computer Science \\
Stanford University\\
Stanford, CA 94305 \\
\texttt{ljkong@stanford.edu} \\
\And
Yun Liao \\
Department of Electrical Engineering \\
Stanford University\\
Stanford, CA 94305 \\
\texttt{yunliao@stanford.edu} \\
}

\begin{document}

\maketitle

\begin{abstract}
Graph neural networks~(GNN) extends deep learning to graph-structure dataset. Similar to Convolutional Neural Networks (CNN) using on image prediction, convolutional and pooling layers are the foundation to success for GNN on graph prediction tasks. In the initial PAN paper \cite{ma2020path} , it uses a path integral based graph neural networks for graph prediction. Specifically, it uses a convolution operation that involves every path linking the message sender and receiver with learnable weights depending on the path length, which corresponds to the maximal entropy random walk. It further generalizes such convolution operation to a new transition matrix called maximal entropy transition (MET). Because the diagonal entries of the MET matrix is directly related to the subgraph centrality, it provide a trial mechanism for pooling based on centrality score. While the initial PAN paper only considers node features. We further extends its capability to handle complex heterogeneous graph including both node and edge features.
\end{abstract}

\section{Introduction}
\subsection{Background}
With the success of applying Convolutional Neural Networks (CNN) on with fix-size 2D image dataset, researchers have dived deeper into how to apply deep learning on graph dataset. \acp{gnn} especially Graph Convolutional Neural Networks (GCN) provides a great framework as baseline. An essential part of GCN is message passing. Message passing not only allow us to encoder richer node feature, but also enables tasks such as node, edge, or even graph prediction. One specific method is to use the graph Laplacian based methods relying on message passing between connected nodes with equal weights across edges. The idea of \ac{grw} defined on graphs is at heart of many graph Laplacian based methods that essentially rely on message passing between directly connected nodes (eg. GCN \cite{kipf2016semi}, GraphSAGE \cite{hamilton2017inductive}, etc.). Although proved effective in many graph-based tasks, the \ac{grw}-based methods inherently suffer from information dilution as paths between nodes branch out. This can pose great difficulty on graph-level interpretation tasks especially when the multi-hop local structures matter as much as the global node attributes. 

Ma, Xuan, et al. proposed a \ac{pan} approach in \cite{ma2020path} that overcomes this drawback by considering every path linking the message sender and receiver as the elemental unit in message passing, which is analogous to Feynman's path integral formulation \cite{feynman2010quantum} extensively used in statistical mechanics and stochastic processes. Similar ideas have been shown effective in link prediction \cite{li2011link} and community detection \cite{ochab2013maximal} tasks. The popular graph attention mechanism \cite{velivckovic2018graph} can also be viewed as a special case of \ac{pan} by restricting the \ac{met} matrix to a particular form. Another stream of related works are the GNN models using multi-scale information and/or higher-order adjacency matrix, such as LanczosNet \cite{liao2019lanczosnet}, N-GCN \cite{abu2020n} and SGC \cite{wu2019simplifying}.

This project re-implements and improves upon the \ac{pan} approach and evaluates its performance on a molecular classification dataset named ogbg-molhiv.

\subsection{Method: \ac{pan} and \acs{hpan}}

The \ac{pan} approach consists two major modules: (1) a \ac{panc} that gives the message passing rule in the graph, and (2) a \ac{panp} that specifies the approach to extract higher-level features of the graph for the final graph-level tasks. 

\begin{figure}
    \centering
    \begin{subfigure}[t]{0.45\textwidth}
        \centering
        \includegraphics[width=\textwidth]{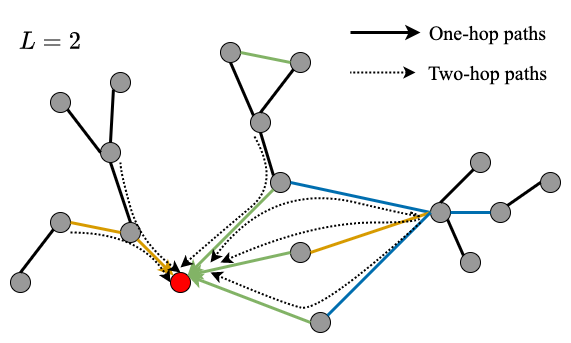}
        \caption{\ac{pan}Conv: Aggregate messages from each path connected to the central node, where the path length is at most $L$. Different colors represent different edge types.}
        \label{fig:pan_conv}
    \end{subfigure}
    \hspace{2em}
    \begin{subfigure}[t]{0.45\textwidth}
        \centering
        \includegraphics[width=\textwidth]{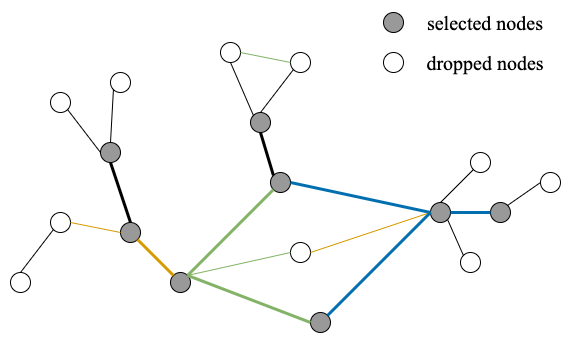}
        \caption{\ac{pan}Pool: obtain the subgraph consisting of $K$ out of $N$ nodes with the highest scores and the edges connecting them.}
        \label{fig:pan_pool}
    \end{subfigure}
    \caption{Visual illustration of the PANConv and PANPool approaches.}
    \label{fig:pan_diagram}
\end{figure}

The \ac{panc} module is illustrated in Fig.~\ref{fig:pan_conv}. Each message passing step aggregates neighborhood information up to $L$ hops away. The message from node $A$ to node $B$ is weighted according to the number of paths between $A$ and $B$ and the lengths of the paths. This can be considered as an extension to the idea of aggregating information from one-hop neighborhood in most of the \ac{gnn} approaches.

The \ac{panp} module computes the score of each node in the graph and selects a subset of nodes with the highest scores. The subgraph induced from the selected nodes is passed to the next layer. All the other nodes and edges are dropped. The \ac{panp} module is illustrated in Fig.~\ref{fig:pan_pool}. 

The original \ac{pan} approach can only deal with node features, and no edge attributes are considered. As an extension to the \ac{pan} approach, this project proposes \ac{hpan} to incorporate edge features in the message passing rule, so that the approach is able to handle heterogeneous graphs. In particular, we add a PanLump layer as shown in figure \ref{fig:pan_net} to sum corresponding node embedding and feature embedding together during message passing.

\subsection{Dataset}
This project uses the ogbg-molhiv dataset, which is adopted from the MoleculeNet, to evaluate the performance of PanConv and hybrid PanPool approach. The ogbg-molhiv dataset contains 41,127 graphs. Each graph represents a molecule, where nodes are atoms, and edges are chemical bonds. The input node features are 9-dimensional, containing atomic number, chirality, etc. The edge features are 3-dimensional, representing bond type, bond stereochemistry, and conjugation, respectively. The task associated with this dataset is to predict whether a target molecule inhibits HIV virus replication or not, which is a binary classification task. ROC-AUC metric will be used for evaluation.

This dataset is selected to evaluate the \ac{pan} and \ac{hpan} algorithms for the following reasons: the key idea of \ac{panc} is to smartly exploit the structural features of the graph in each layer of convolution, while on the other hand, \ac{panp} provides a way to combine the node features with the local graph structure. This can be very useful in understanding graphs representing molecules, where the structural information can be essential. Moreover, the output of the selected algorithm can be naturally interpreted as a graph-level representation, which makes it convenient for a graph-level classification task.

\section{Existing Method}
This section details the \ac{pan} approach, which is \textbf{NOT YET in the OGB leaderboard}. The \ac{pan} approach consists two major components: \ac{panc} as the convolution operator and \ac{panp} as the pooling operator. 

\subsection{\ac{panc}}
\begin{figure}
    \centering
    \includegraphics[width=0.7\textwidth]{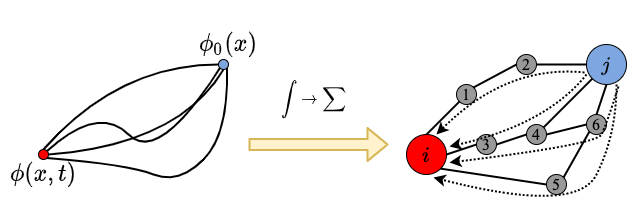}
    \caption{Schematic analogy between the original path integral formulation in continuous space (left) and the discrete version for a graph (right).}
    \label{fig:pan_idea}
\end{figure}

The core of \ac{panc} is a discrete analogy of Feynman’s path integral formulation \cite{feynman2010quantum} that can be applied to a graph. The analogy is illustrated in Fig.~\ref{fig:pan_idea}. In the original path integral formulation, the probability amplitude $\phi(\bm{x},t)$ is influenced by the surrounding field, where the contribution from $\phi_0(\bm{x})$ is computed by summing over the influences (denoted by $e^{iS[\bm{x}, \dot{\bm{x}}]}$) from all paths connecting itself and $\phi(\bm{x},t)$:
\begin{equation}
    \phi(\bm{x},t) = \frac{1}{Z} \int \phi_0(\bm{x}) \int e^{iS[\bm{x}, \dot{\bm{x}}]} \mathcal{D}(\bm{x}),
\end{equation}
where $Z$ is the partition function.

A graph can be viewed as a discrete version of a continuous field. The path integral formulation is generalized to a graph setting in \cite{ma2020path} by replacing the integral over paths to a discrete sum over all possible paths in the graph, and the integral of Lagrangian $e^{iS[\bm{x}, \dot{\bm{x}}]}$ to a sum over Boltzmann's factor. The analogous path integral formulation on graph $G(V,E)$ then becomes
\begin{equation}\label{eq:pan_on_graph_general}
    \phi_i = \frac{1}{Z_i} \sum_{j \in V} \phi_j \sum_{\{\bm{l}|l_0 = i, l_{|\bm{l}| }= j\}} e^{-\frac{E[\bm{l}]}{T}} \stackrel{(*)}{=} \frac{1}{Z_i} \sum_{l = 0}^{\infty} e^{-\frac{E(l)}{T}} \sum_{j \in V} g(i,j;l) \phi_j,
\end{equation}
where $\phi_i$ denotes the feature/message at node $i$, $\bm{l} = (l_0, l_1, \ldots, l_{|\bm{l}|})$ is a path connecting node $i$ and node $j$, $E[\bm{l}]$ represents the fictitious energy w.r.t. path $\bm{l}$, and $T$ is the fictitious energy. $Z_i$ is the partition function for node $i$. Equation $(*)$ holds under the assumption that the fictitious energy only depends on the path length $l$, in which $g(i,j;l)$ denotes the number of length-$l$ paths between nodes $i$ and $j$. Presumably, the energy $E(l)$ is an increasing function of $l$. Note that $g(i,j;l)$ can be computed efficiently from the graph's adjacency matrix $A$ as $(A^l)_{ij}$. To make the computation tractable, a cutoff maximal path length $L$ is applied to \eqref{eq:pan_on_graph_general}, and the path integral formulation is simplified as
\begin{equation}\label{eq:pan_on_graph}
    \phi_i = \frac{1}{Z_i} \sum_{l = 0}^{L} e^{-\frac{E(l)}{T}} \sum_{j \in V} (A^l)_{ij} \phi_j.
\end{equation}
The above expression can be written in a compact form by defining the \ac{met} matrix
\begin{equation}\label{eq:panconv_met}
    M = Z^{-1} \sum_{l = 0}^L e^{-\frac{E(l)}{T}}A^l,
\end{equation}
where $Z = {\rm diag}(Z_i)$. The name comes from the fact that it realizes maximal entropy under the microcanonical ensemble. Based on the \ac{met} matrix, a convolutional layer can be defined as
\begin{equation}\label{eq:panconv_vanilla}
    X^{(h+1)} = M^{(h)}X^{(h)}W^{(h)},
\end{equation}
where $h$ is the layer index and $W^{(h)}$ is the trainable weight. A \ac{panc} module further improves over \eqref{eq:panconv_vanilla} by applying symmetric normalization instead of $Z^{-1}$. The final definition of \ac{panc} is
\begin{equation}
    X^{(h+1)} = Z^{-1/2} \sum_{l = 0}^L e^{-\frac{E(l)}{T}}A^l Z^{-1/2} X^{(h)} W^{(h)}.
\end{equation}

\subsection{\ac{panp}}
The diagonal element $M_ii$ of the \ac{met} matrix resembles the subgraph centrality defined by $\sum_{l = 1}^{\infty} (A^l)_{ii}$ for node $i$, so ${\rm diag}(M)$ provides a natural characterization of the nodes' importance. On the other hand, the global importance can be represented by, but not limited to, the strength of the message $X$ itself. \ac{panp} projects feature $X \in \mathbb{R}^{|V| \times d}$ by a trainable parameter vector $p \in \mathbb{R}^{d}$ and combines it with ${\rm diag}(M)$ to obtain a score vector
\begin{equation}
    {\rm score} = Xp + \beta {\rm diag}(M)
\end{equation}
Here, $\beta \in \mathbb{R}$ controls the emphasis on these two potentially competing factors. \ac{pan}Pool then selects a fraction of the nodes ranked by this score (number denoted by $K$), and outputs the pooled feature array $\tilde{X} \in \mathbb{R}^{K \times d}$ and the corresponding adjacency matrix $\tilde{A} \in \mathbb{R}^{K \times K}$.

The feature array of the final layer of the stacked ``\ac{pan}Conv + \ac{pan}Pool'' structure is averaged and then fed into the task-oriented layers, which can be trained against any ground truth label through specific loss functions.

\section{Method}
This section elaborates the modifications over the original \ac{pan} approach that could bring improvements on the classification task in the dataset of interest.

\subsection{HPAN with edge features}

The original \ac{pan} approach can only handle homogeneous graph without edge attributes, and only node attributes are being propagated in the graph. However, the dataset of interest (ogbg-molhiv) includes edge attributes, direct application of the \ac{pan} approach to this dataset would simply ignore the edge attributes and treat the edges as if they are homogeneous, which would lose information and potentially result in inferior classification performance (as shown by the experimental results in Section~\ref{sec:experiments}.

Our method named \ac{hpan} proposes to incorporate the edge features in the \ac{pan} approach by introducing an additional module named PANLump.

\begin{figure}
    \centering
    \includegraphics[width=0.7\textwidth]{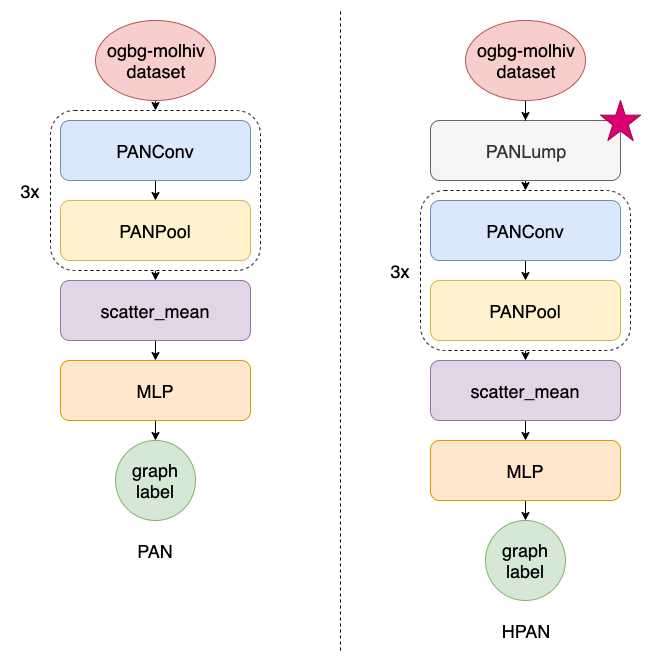}
    \caption{Comparison of a \ac{pan} structure and the corresponding \ac{hpan} structure.}
    \label{fig:pan_net}
\end{figure}

The \ac{pan}Lump module is added before the stack of ``\ac{panc} + \ac{panp}'' layers as shown in Fig.~\ref{fig:pan_net}. This module adds transformed edge features to the incident node features. The \ac{pan}Lump module works as follows. First, apply \texttt{atom\_encoder} to all nodes and \texttt{edge\_encoder} to all edges, where the embedding dimensions of both encoders match each other. Then, each edge $(u,v)$ shares its embedding $\bm{h}(u,v)$ to both of its vertices, and the incident node embeddings $\bm{X}(u)$ (as well as $\bm{X}(v)$) update as
\begin{equation}
    \bm{X}'(u) \leftarrow \mathbf{MLP}[(1 + \epsilon)\bm{X}(u) + \mathbf{AGG}\{\bm{h}(u,v),~\forall v \in \mathcal{N}(u)\})].
\end{equation}
The output of the \ac{pan}Lump module is the updated node features that incorporate its surrounding edge attributes. This output is then fed into the following ``\ac{panc} + \ac{panp}'' structure, which assumes homogeneous edges.

\subsection{Loss function}
The ogbg-molhiv dataset is highly skewed in the sense that it contains 39,684 samples (not inhabit HIV) and only 1,443 positive samples (inhibit HIV). To make sure that the model does not learn trivially to always output 0, a weighted binary cross entropy loss is adopted in training. In particular, the loss function puts larger weight on positive sample than on negative ones:
\begin{equation}
    loss = \frac{1}{|\mathcal{D}|}\sum_{n = 1}^{|\mathcal{D}|} \left[\alpha y^{(n)} \log\left(\hat{y}^{(n)}\right) + \left(1-y^{(n)}\right) \log \left(1-\hat{y}^{(n)}\right)\right],
\end{equation}
where $\hat{y}^{(n)}$ is the predicted probability of the $n$-th graph sample has positive label, and the weight $\alpha \ge 1$.

\section{Experiments and Discussions}\label{sec:experiments}

\begin{table}[t]
    \centering
    \caption{Comparison of ROC-AUC scores of ogbg-molhiv dataset}
    \begin{tabular}{||c c c c c c||} 
     \hline
     \textbf{Model} & \textbf{Add. Feat.} & \textbf{Virtual Node} & \textbf{ROC-AUC Val} & \textbf{ROC-AUC Test} & \textbf{\#Params} \\ [0.5ex] 
     \hline\hline
     GCN & No & Yes & 83.73 $\pm$ 0.78 & 74.18 $\pm$ 1.22 & 1,978,801\\ 
     \hline
     GCN & Yes & No & 82.04 $\pm$ 1.41 & 76.06 $\pm$ 0.97 & 527,701\\
     \hline
     GCN & Yes & Yes & 83.84 $\pm$ 0.91 & 75.99 $\pm$ 1.19 & 1,978,801\\ 
     \hline
     GIN & No & Yes & 84.1 $\pm$ 1.05 & 75.2 $\pm$ 1.30 & 3,336,306 \\
     \hline
     GIN & Yes & No & 82.32 $\pm$ 0.90 & 75.58 $\pm$ 1.40 & 1,885,206 \\
     \hline
     GIN & Yes & Yes & \textbf{84.79} $\pm$ 0.68 & \textbf{77.07} $\pm$ 1.49 & 3,336,306\\ \hline \hline
     \textbf{PAN} & Yes & No & 70.17 $\pm$ 0.29 & 73.06 $\pm$ 0.49 & \textbf{26,843} \\
     \hline
     \textbf{HPAN} & Yes & No & 82.27 $\pm$ 0.44 & 76.76 $\pm$ 0.41 & \textbf{43,676} \\
     \hline
    \end{tabular}
    \label{table:result}
\end{table}

Both PAN and HPAN are implemented and evaluated on the OGB ogbg-molhiv dataset \cite{hu2020open}. The evaluation metric is the ROC-AUC score. The performance is compared to the benchmarks given in \cite{hu2020open}. In particular, the benchmark includes GCN and GIN with or without using additional features and virtual nodes. The results are shown in Table~\ref{table:result}. 

In the experiments, both \ac{pan} and \ac{hpan} use 3 layers of \ac{panc} and \ac{panp} followed by a mean pooling module to extract a graph-level representation. The output representation is then fed into a two-layer MLP of size $\{\text{emb\_dim}, \text{emb\_dim}/2, 1\}$ and ReLU activation. Cutoff length $L=3$ for \ac{panc}1, and $L=2$ for \ac{panc}2 and \ac{panc}3. At each \ac{panp} layer, 80\% of the nodes are selected according to the score function, i.e., $|V|^{(h+1)} = K = 0.8 |V|^{(h)}$. The MLP in the \ac{pan}Lump module has size $\{\text{emb\_dim}, \text{emb\_dim}*2, \text{emb\_dim}\}$ with BatchNorm and ReLU activation. The sum aggregation is adopted as $\mathbf{AGG}(\cdot)$ in \ac{pan}Lump. The embedding dimension is set to 64, and the weight in loss function is set to $\alpha = 5.0$ in \ac{pan} and $\alpha = 10.0$ in \ac{hpan}. The ROC-AUC scores and variations are calculated based on 10 independent runs.

As shown in the table, the original \ac{pan} without considering edge features does not show very satisfying results. Instead, the modified version \ac{hpan} shows comparable performance to the benchmarks. The deviation, on the other hand, is significantly reduced compared to the benchmarks, which may be due to the consideration of higher order paths. Besides, since each \ac{panc} layer considers information in up to $L$-hop neighborhoods, fewer layers are needed to extract the information from the entire graph. Therefore, \ac{pan} and \ac{hpan} can be extremely light-weighted. This is verified in Table~\ref{table:result} in the sense that the number of trainable parameters is several orders of magnitude smaller than the parameters needed for the benchmarks. 

\section{Conclusion}
The \ac{pan} approach, which takes multi-hop message passing into account in each graph convolutional layer, was implemented and applied to the graph classification task for the ogbg-molhiv dataset. A \ac{hpan} method was proposed as an extension to the original \ac{pan} approach that enables the model to incorporate edge attributes. Experimental results showed that the original \ac{pan} approach gave inferior performance compared to the existing benchmarks due to the ignorance of edge features, while the modified \ac{hpan} method achieved comparable performance to the benchmarks with a extremely small model. Besides, \ac{pan} and \ac{hpan} also showed significantly smaller variance in performance across independent runs. In this perspective, PAN provides a more efficient and stable learning model for the graph classification task. Future works would include smarter ways to incorporate edge features in each layer of \ac{panc} or \ac{panp} and theoretical comparison between wide \acp{pan} (i.e., large $L$) and deep GNN.

{\small
\bibliographystyle{unsrt}
\bibliography{bibliography}

\begin{thebibliography}{10}

\bibitem{ma2020path}
Zheng Ma, Junyu Xuan, Yu~Guang Wang, Ming Li, and Pietro Li{\`o}.
\newblock Path integral based convolution and pooling for graph neural
  networks.
\newblock {\em arXiv preprint arXiv:2006.16811}, 2020.

\bibitem{kipf2016semi}
Thomas~N Kipf and Max Welling.
\newblock Semi-supervised classification with graph convolutional networks.
\newblock In {\em International Conference on Learning Representations (ICLR)},
  2017.

\bibitem{hamilton2017inductive}
William~L Hamilton, Rex Ying, and Jure Leskovec.
\newblock Inductive representation learning on large graphs.
\newblock In {\em Proceedings of the 31st International Conference on Neural
  Information Processing Systems}, pages 1025--1035, 2017.

\bibitem{feynman2010quantum}
Richard~P Feynman, Albert~R Hibbs, and Daniel~F Styer.
\newblock {\em Quantum mechanics and path integrals}.
\newblock Courier Corporation, 2010.

\bibitem{li2011link}
Rong-Hua Li, Jeffrey~Xu Yu, and Jianquan Liu.
\newblock Link prediction: the power of maximal entropy random walk.
\newblock In {\em Proceedings of the 20th ACM international conference on
  Information and knowledge management}, pages 1147--1156, 2011.

\bibitem{ochab2013maximal}
JK~Ochab and Zdzis{\l}aw Burda.
\newblock Maximal entropy random walk in community detection.
\newblock {\em The European Physical Journal Special Topics}, 216(1):73--81,
  2013.

\bibitem{velivckovic2018graph}
Petar Veli{\v{c}}kovi{\'c}, Guillem Cucurull, Arantxa Casanova, Adriana Romero,
  Pietro Lio, and Yoshua Bengio.
\newblock Graph attention networks.
\newblock In {\em International Conference on Learning Representations (ICLR)},
  2018.

\bibitem{liao2019lanczosnet}
Renjie Liao, Zhizhen Zhao, Raquel Urtasun, and Richard~S Zemel.
\newblock Lanczosnet: Multi-scale deep graph convolutional networks.
\newblock {\em arXiv preprint arXiv:1901.01484}, 2019.

\bibitem{abu2020n}
Sami Abu-El-Haija, Amol Kapoor, Bryan Perozzi, and Joonseok Lee.
\newblock N-gcn: Multi-scale graph convolution for semi-supervised node
  classification.
\newblock In {\em uncertainty in artificial intelligence}, pages 841--851.
  PMLR, 2020.

\bibitem{wu2019simplifying}
Felix Wu, Amauri Souza, Tianyi Zhang, Christopher Fifty, Tao Yu, and Kilian
  Weinberger.
\newblock Simplifying graph convolutional networks.
\newblock In {\em International conference on machine learning}, pages
  6861--6871. PMLR, 2019.

\bibitem{hu2020open}
Weihua Hu, Matthias Fey, Marinka Zitnik, Yuxiao Dong, Hongyu Ren, Bowen Liu,
  Michele Catasta, and Jure Leskovec.
\newblock Open graph benchmark: Datasets for machine learning on graphs.
\newblock {\em arXiv preprint arXiv:2005.00687}, 2020.

\end{thebibliography}
}

\end{document}